\documentclass[11pt,letterpaper]{llncs}
\usepackage{amssymb}
\usepackage{marvosym}
\usepackage{fancyhdr}

\usepackage{geometry}
\usepackage{graphicx}
\usepackage{amsmath}
\usepackage{bm}
\usepackage{comment}
\usepackage[lined,ruled]{algorithm2e}
\usepackage{float}
\usepackage{bbm}
\usepackage{amsfonts}
\usepackage{caption}
\usepackage{graphicx}
\usepackage{pdflscape}
\usepackage{booktabs, makecell}
\newcommand{\ra}[1]{\renewcommand{\arraystretch}{#1}} 
\usepackage{multirow}
\usepackage[font=scriptsize]{caption}
\usepackage[hyphens]{url} 
\usepackage{hyperref}
\hypersetup{
    urlcolor  = blue,
	colorlinks = true,
	citecolor = blue, 
	linkcolor = blue 
}

\DeclareMathOperator{\ave}{ave}
\DeclareMathOperator{\quantile}{quantile}
\DeclareMathOperator{\AD}{AD}

\newcommand{\keywords}[1]{\par\addvspace\baselineskip
\noindent\keywordname\enspace\ignorespaces#1}

\fancypagestyle{firstpage}{\fancyhf{}
\fancyhead[C]{\small{
}}
\fancyfoot[L]{DOI: ...
}
\rfoot{\thepage}
}

\begin{document}


\title{\LARGE{DTOR: Decision Tree Outlier Regressor to explain anomalies}\thanks{The views and opinions expressed are those of the authors and do not necessarily reflect the views of Intesa Sanpaolo, its affiliates or its employees.}}


%
%

\author{\large{Riccardo Crupi\inst{1}\orcidID{0009-0005-6714-5161} \and
Daniele Regoli\inst{1}\orcidID{0000-0003-2711-8343} \and
Alessandro Damiano Sabatino\inst{1}\orcidID{0000-0002-1336-2057} \and
Immacolata Marano\inst{2} \and
Massimiliano Brinis\inst{2} \and
Luca Albertazzi\inst{2} \and
Andrea Cirillo\inst{2} \and
Andrea Claudio Cosentini\inst{1}}
}
\authorrunning{R. Crupi et al.}
%
\institute{\large{Data Science \& Artificial Intelligence, Intesa Sanpaolo, Italy \and
Audit Data \& Advanced Analytics, Intesa Sanpaolo, Italy}
\email{<name>.<surname>@intesasanpaolo.com}}

%


%
%


\maketitle

\thispagestyle{firstpage}

\begin{abstract}
Explaining outliers occurrence and mechanism of their occurrence can be extremely important in a variety of domains. Malfunctions, frauds, threats, in addition to being correctly identified, oftentimes need a valid explanation in order to effectively perform actionable counteracts. The ever more widespread use of sophisticated Machine Learning (ML) approach to identify anomalies make such explanations more challenging. We present the Decision Tree Outlier Regressor (DTOR), a technique for producing rule-based explanations for individual data points by estimating anomaly scores generated by an anomaly detection model. This is accomplished by first applying a decision tree regressor, which computes the estimation score, and then extracting the relative path associated with the data point score. Our results demonstrate the robustness of DTOR even in datasets with a large number of features. Additionally, in contrast to other rule-based approaches, the generated rules are consistently satisfied by the points to be explained. Furthermore, our evaluation metrics indicate comparable performance to Anchors in outlier explanation tasks, with reduced execution time.

\end{abstract}


\section{Introduction}

\subsection{Anomaly detection in the Banking sector - Internal Audit activity}\label{sec:intro_ad}




The objectives of internal audit in the banking system are multifaceted and crucial for ensuring the integrity and efficiency of operations. Internal auditors aim to assess and evaluate the effectiveness of internal controls, risk management processes and compliance with regulatory requirements. They also strive to identify and mitigate potential risks, detect fraud, and provide recommendations for improvement \cite{nonnenmacher2021unsupervised}-\cite{basile2024disambiguation}. 

In this context, anomaly detection techniques play a significant role in identifying atypicalities within the populations of data being analyzed for audit purposes. By leveraging advanced algorithms, these techniques can effectively identify patterns and outliers that deviate from the norm. The use of anomaly detection algorithms is particularly valuable as a mean to define a sample of records to be reviewed as they provide an anomaly score, allowing for the ranking of records based on their level of abnormality, rather than a simple binary classification. 

Nevertheless, to ensure that these techniques can be effectively employed in an internal audit function, the ability to explain why certain records are considered atypical compared to others is of utmost importance, since the sample extracted from the total population is subsequently reviewed in detail by internal auditors who may not possess extensive expertise in data analytics matters. The explainability aspect of anomaly detection algorithms enables internal auditors to understand the underlying reasons behind the anomalies, facilitating their decision-making process and enhancing their ability to identify potential risks or fraudulent activities. This empowers auditors to effectively address and resolve issues, ultimately contributing to the overall integrity and security of the banking system.


Several anomaly detection techniques have been developed and employed in the banking sector, each offering distinct advantages and trade-offs. Among these techniques, three notable approaches, which leverage diverse mathematical principles, have gained prominence (\cite{zhao2019pyod}-\cite{kumar2019detecting}): 

\begin{enumerate}
    \item Isolation Forest~\cite{liu2008isolation}: a tree-based ensemble method designed to isolate anomalies by recursively partitioning data points until they are uniquely isolated. Anomalies are identified as instances that require significantly fewer partitions to isolate, distinguishing them from normal data points efficiently.

    \item One-Class Support Vector Machine (SVM)~\cite{scholkopf1999support}: a variant of the Support Vector Machine algorithm
    tailored for anomaly detection. It learns a hyperplane that encloses the majority of data points in a high-dimensional space, effectively defining the "normal" region. Instances lying outside this region are considered anomalies, enabling the detection of outliers in a dataset with limited or no labeled anomalies.

    \item Gaussian Mixture Models (GMM)~\cite{reynolds2009gaussian}: GMM models data distribution as a combination of Gaussian distributions. Each Gaussian component represents a cluster of similar data points. Anomalies are detected by identifying data points with low probabilities, indicating they are unlikely to have been generated by the model. 

\end{enumerate}

\subsection{eXplainable AI for anomaly detection}

In addition to developing effective anomaly detection techniques, it is imperative to understand and interpret the decisions made by these models, especially in banking where transparency and accountability are fundamental. XAI (eXplainable Artificial Intelligence) techniques plays a pivotal role in elucidating the inner workings of complex models, enabling stakeholders to comprehend the factors driving model predictions and actions.

One prominent XAI technique is SHAP (SHapley Additive exPlanations)~\cite{lundberg2020local}, which operates in a model-agnostic manner. SHAP provides insights into the contributions of individual features towards model predictions by estimating the Shapley values, a concept derived from Cooperative Game Theory~\cite{shapley1951notes}. By quantifying the impact of each feature on the model output, SHAP is capable of helping researchers to understand how the models make decisions.

In contrast, the recently proposed DIFFI (Depth-based Feature Importance of Isolation Forest)~\cite{CARLETTI2023105730} offers a specialized approach to explain the Isolation Forest algorithm specifically tailored for anomaly detection tasks. The Isolation Forest, renowned for its efficacy in anomaly detection, often is affected by a lack of interpretability due to the inherent randomness in its decision-making process. DIFFI addresses this challenge by defining feature importance scores at both global and local levels, providing valuable insights into the anomaly detection process. Moreover, DIFFI introduces a procedure for unsupervised feature selection, enhancing the interpretability of Isolation Forest models while maintaining computational efficiency.

However, while XAI techniques based on feature importance, such as SHAP, provide valuable insights into the relative importance of individual features, their interpretability may be limited, especially in complex models or for high-dimensional datasets. The reliance on feature importance scores alone may not sufficiently capture the underlying decision-making process, leading to challenges in understanding the model's behavior comprehensively.

To address this limitation, rule-based XAI techniques, such as Anchors~\cite{Ribeiro_Singh_Guestrin_2018}, offer an alternative approach to model interpretation. Anchors approach generates human-interpretable rules that explain model predictions by identifying conditions under which the predictions hold true. By formulating rules that encapsulate the model's decision logic, Anchors enhance interpretability and facilitates trust in AI-driven decisions. 

Although Anchors excel in providing transparent and comprehensible explanations, it is not explicitly designed for anomaly detection tasks, for instance is not suited for a regression task. Herein lies the motivation for our work. We recognize the critical need for interpretable XAI techniques tailored specifically for anomaly detection in the banking sector. Hence, we introduce a novel model-agnostic XAI framework explicitly suited for anomaly detection and rule-based interpretation. 
In related literature, approaches like LORE (LOcal Rule-based Explanations) \cite{guidotti2018local} provides decision rules by training a decision tree on local data points around the instance in question. Similarly, RuleXAI \cite{macha2022rulexai} offers rule-based explanations for regression and classification models. Other approaches involve reasoning by identifying the nearest sample for which the label is changed, known as counterfactuals \cite{crupi2022counterfactual}, although this aspect is not further analyzed in our work.

Our approach aims to bridge the gap between interpretability and effectiveness in anomaly detection by providing human-understandable rules that elucidate the reasons behind anomalous predictions. By leveraging rule-based explanations, our XAI framework ensures that the decision-making process of anomaly detection models is transparent and accessible to both data scientists and domain experts, including our colleagues in the banking industry.

In the subsequent sections of this paper, we delve into the development and evaluation of our model-agnostic XAI framework, highlighting its efficacy in enhancing the interpretability of anomaly detection models and fostering collaboration and trust among stakeholders in the banking sector.

\section{Method}
Drawing inspiration from the principles of the Isolation Forest algorithm, our novel XAI method capitalizes on the notion of isolating anomalies with minimal cuts in the feature space. To provide transparent explanations for anomaly detection decisions, we employ decision tree regressors. In our approach, a decision tree regressor is trained to learn the anomaly scores associated with each data point generated by the Anomaly Detector (AD). Notably, we introduce a weighted loss function during training, assigning a significantly higher weight to the data point under scrutiny. This weighting scheme ensures that the decision tree regressor prioritizes accurate estimation of the anomaly score for the target data point, enhancing the interpretability and reliability of the local explanation. Once the decision tree is trained, it is sufficient to extract the path followed by the datapoint to be explained to actually provide an interpretable rule for its anomaly score~(\autoref{alg:DTOR.explain}). 

In accordance with Ribeiro et al. \cite{Ribeiro_Singh_Guestrin_2018}, we establish notation for a black-box classifier $f$, an instance $x$, and a data distribution $\mathcal{D}$\footnote{In this work we can consider this distribution as the empirical distribution of the training set.}. A condition rule $A_x$ comprises a set of feature predicates on $x$ aimed at achieving a level of precision denoted as $\text{precision}(A_x)$, defined as:
\begin{equation}
    \text{precision}(A_x) = \mathbb{E}_{\mathcal{D}(z\mid A_x)} \left[ \mathbbm{1}_{f(x) = f(z)} \right]. \label{eq:precision}
\end{equation}
Namely, precision quantifies to what extent rule $A_x$ is sufficient to be given $f(x)$ as anomaly signal. 

Coverage is quantified as the proportion of data points in the dataset that effectively satisfy rule $A_x$, given by:
\begin{equation}
   \text{coverage}(A_x) = \mathbb{E}_{\mathcal{D}(z)}\left[A_x(z)\right], \label{eq:coverage}
\end{equation}
where it is understood that:
\begin{equation}
    A_x(z) = \left\{\begin{aligned} 1 && \text{if}\ z\ \text{satisfies the rule}\ A_x, \\ 0 && \text{otherwise.}\end{aligned}\right.
\end{equation}
For instance, if the rule $A_x$ is defined as $A_x(z) = (z_1 > 1) \; AND \; (z_2 > 5)$, the candidate sample $\hat{z}=[3, 10]$ satisfy the rule, therefore $A_x(\hat{z})=1$. 

Furthermore, while seemingly trivial, assessing the validity of a condition rule as an explanation for the instance $x$ being examined is crucial:
\begin{equation}
   \text{validity}(A_x) = A_x(x). \label{eq:validity}
\end{equation}

\begin{algorithm}
    \SetStartEndCondition{ }{}{}%
    \SetKwProg{Fn}{def}{\string:}{}
    \AlgoDontDisplayBlockMarkers\SetAlgoNoEnd\SetAlgoNoLine%
    \SetKwInOut{Input}{input}\SetKwInOut{Output}{output}
    \Fn{explain\_instance{}}{
    
        \Input{
            $(x_{\text{expl}}, \hat y_{\text{expl}})$: an instance and its anomaly score to be explained\;\\ 
            $(X_{\text{train}}, \hat{y}_\text{train})$: a training dataset and its anomaly scores\;\\
            $\beta$: relative weight given to $x_\text{expl}$\;\\
            \emph{h-args}: set of hyperparameters for the Decision Tree Regressor\;
        }
        \Output{a set of rules explaining $(x_{\text{expl}}, \hat y_{\text{expl}})$}
    
        $N \leftarrow$ len($X_\text{train}$)\; 
        DT $\leftarrow$ DecisionTreeRegressor(\emph{h-args})\; 
        \tcc{extend the training set with the instance to be explained}
        $X \leftarrow$ concatenate $X_{\text{train}}$ with $x_{\text{expl}}$\;
        $y \leftarrow$ concatenate $\hat{y}_\text{train}$ with $\hat{y}_\text{expl}$\;
        \tcc{define Decision Tree loss weights assigning more importance to the instance to be explained}
        $w\leftarrow $ concatenate $\bm{1}_N$ with $\beta$\;
        \tcc{fit the decision tree to the `weighted' training set}
        DT.fit($(X, y)$, sample\_wights=$w$)\;
        \tcc{extract the path of the fitted DT followed by $x_\text{expl}$}
        rules $\leftarrow$ extract\_path(DT, $x_\text{expl}$)\;
        \textbf{return} rules

    }

\caption{DTOR method to generate explanations for a given instance.}
\label{alg:DTOR.explain}
\end{algorithm}
 
Notice that in order to compute precision and coverage from eqs.~\eqref{eq:precision}-\eqref{eq:coverage} we need a prescription to sample from $\mathcal{D} \mid A_x$ and $\mathcal{D}$, respectively. While the latter is more or less straightforward, since we can easily estimate unconditional distribution by sampling from a given ``training'' set, the former is more subtle for a couple of reasons. On the one hand, estimating $\mathcal{D} \mid A_x$ as the empirical conditional distribution in the training set is often a poor choice, since the number of instances satisfying $A_x$ may be very small. On the other hand, estimating $\mathcal{D} \mid A_x$ as the conditional distribution may also be sub-optimal for the purpose of computing the effectiveness of $A_x$ as an explanation of $f(x)$, due to correlations with variables not comprised as predicate in the rule $A_x$.  

For tabular data, referring to the original Anchors paper~\cite{Ribeiro_Singh_Guestrin_2018}, the authors propose to estimate $\mathcal{D}\mid A_x$ by sampling from $\mathcal{D}$ (the training set), and then fixing the variables comprised in $A_x$ such that the sampled instances actually satisfy $A_x$. This approach completely breaks correlation within $\mathcal{D}$ between variables in $A_x$ and variables not in $A_x$. It actually results in sampling also out-of-distribution instances, raising doubts that the computation of $f$ on such datapoints outputs meaningful results. 

To compute precision for DTOR, we propose a slight modification of this approach, that tries to preserve at least some level of correlations found in $\mathcal{D}$ between $A_x$ variables and non-$A_x$ variables. Namely, we first extract from the training set the instances that do satisfy $A_x$, if any. Then we select instances that satisfy all but one predicates in $A_x$, and fix the value of the non-satisfied variable so that those instances satisfy all predicates in $A_x$. We keep selecting instances from the training set, each time lowering by one the number of predicates in $A_x$ that are satisfied. We go on until the number of collected samples reaches a pre-defined hyperparameter integer $N$ denoting the size of the estimation set. If $N$ is not reached at the end of this procedure, the residual instances are sampled exactly as by the approach used by Anchors. 

The pseudocode for the describred procedure is outlined in \autoref{alg:DTOR.precision}.\footnote{for an implementation consisting in both generation of the rule and computing the precision refer to the method \texttt{explain\_instances\_with\_precision()} in \url{https://github.com/rcrupiISP/DTOR/blob/main/dtor.py}.}

\begin{algorithm}
    \SetStartEndCondition{ }{}{}%
    \SetKwProg{Fn}{def}{\string:}{}
    \SetKwIF{If}{ElseIf}{Else}{if}{:}{elif}{else:}{}%
    \SetKw{KwTo}{in}
    \SetKwFor{For}{for}{\string:}{}%
    \AlgoDontDisplayBlockMarkers\SetAlgoNoEnd\SetAlgoNoLine%
    \SetKwInOut{Input}{input}\SetKwInOut{Output}{output}
    \Fn{compute\_precision{}}{
    
        \Input{
            $(x_{\text{expl}}, \hat y_{\text{expl}})$: an instance and its anomaly score to be explained\;\\
            $(X_{\text{train}}, \hat{y}_\text{train})$: a training dataset and its anomaly scores\;\\
            N\_gen: number of synthetic samples to generate\;\\
            AD: the anomaly detector model, outputs a score\;\\
            $t$: threshold for the anomaly score\;\\
            $A$: DTOR rule explaining $(x_{\text{expl}}, \hat y_{\text{expl}})$ as by \autoref{alg:DTOR.explain}\;
        }
        \Output{precision score for the rule $A$ in explaining $(x_{\text{expl}}, \hat y_{\text{expl}})$}
    $X \leftarrow$ concatenate $X_{\text{train}}$ with $x_{\text{expl}}$\;
    $y \leftarrow$ concatenate $\hat{y}_\text{train}$ with $\hat{y}_\text{expl}$\;
    value-grid $\leftarrow$ $\{\}$ \tcc*{initialize the dictionary containing the values per each feature of the rule $A$}
    ids-cond = [\ ] \tcc*{initialize the list containing the indices that satisfy at least one sub-rule $A[k]$}
    \tcc{for loop per each feature in the rule. An example of the rule $A$ could be: "feature\_1 > 5 AND feature\_4 < 0"}
    \For{$k$ in features in rule $A$}{
        \tcc{select the samples that satisfy the partial rule for the specific feature, e.g., "feature\_1 > 5". Note that there can be multiple condition on the same feature can appear, in this case all should go in AND condition}    
        $X_\text{cond} \leftarrow$ elements of $X$ satisfying $A[k]$\;
        ids-cond $\leftarrow$ concatenate ids-cond with the ids of $X_\text{cond}$\;
        \tcc{store the values compatible with sub-rule $A[k]$}
        \If{$k$ is categorical}{
            value-grid$[k] \leftarrow$ unique value of feautre $k$ in $X_\text{cond}$\;
        }
        \Else{
            value-grid$[k] \leftarrow [\min(X_\text{cond}[k]), \quantile_{1/4}(X_\text{cond}[k]), \quantile_{1/2}(X_\text{cond}[k]), \ave(X_\text{cond}[k])$,
            $\quantile_{3/4}(X_\text{cond}[k]), \max(X_\text{cond}[k])]$\;
        }
    }
   \tcc{Give precedence to ids that satisfy multiple sub-rules in $A$}
   ids-cond $\leftarrow$ sort ids-cond by frequency\;    
   \If{length of ids-cond $< N_\text{gen}$}{
        ids-cond $\leftarrow$ concatenate ids-cond with random ids of $X$ to fill the $N_\text{gen}$ samples\;
   }
   $X_\text{synth} \leftarrow$ first $N_\text{gen}$ rows of  $X[\text{ids-cond}]$\;

    \tcc{fix all values of predicates in $A$ of $X_\text{synth}$ so that all samples in $X_\text{synth}$ satisfy rule $A$}
   \For{$k$ in features in rule $A$}{
        $X_\text{synth}[k] \leftarrow$ sample $N_\text{gen}$ times from value-grid$[k]$\;
    }
    \If(\tcc{If the sample to explain is an outlier}){$\hat y_{\text{expl}} < t$}{
        $\text{precision} = \ave\left(\AD(X_\text{synth}) < t\right)$\;
    }
    \Else{
        $\text{precision} = \ave\left(\AD(X_\text{synth}) >= t\right)$\;
    }
    \textbf{return} precision
    }
\caption{DTOR method to generate precision sampling from a generated synthetic dataset. $A[k]$ denotes the $k$-th predicate of rule $A$.}
\label{alg:DTOR.precision}
\end{algorithm}

\begin{figure}[!hbt]
\centering
\includegraphics[width=1\textwidth]{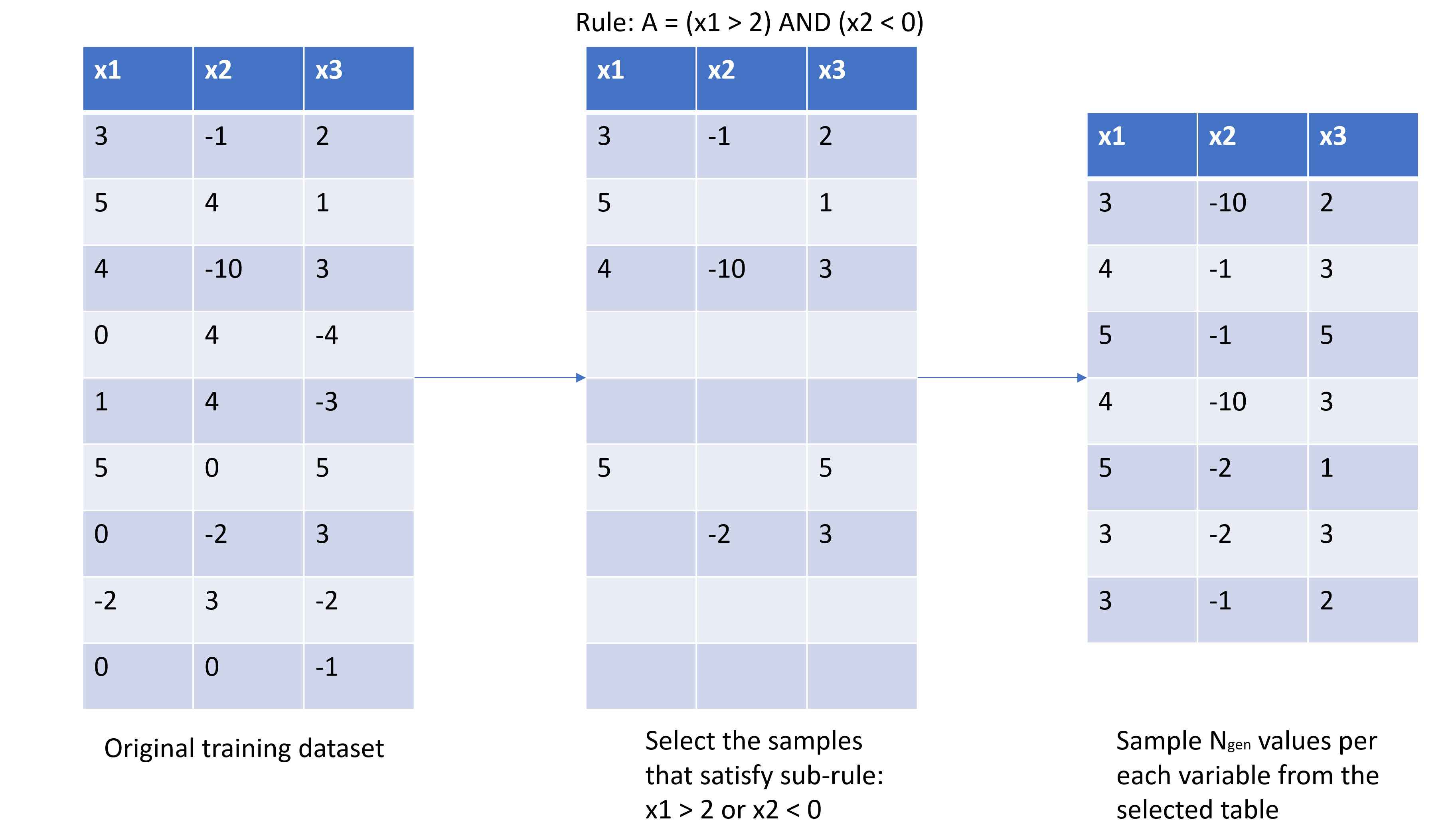}
\caption{A simplified illustration of synthetic data generation is presented. Initially, samples from the original dataset are selected based on sub-rules (e.g., $x_1>2$ or $x_2<0$ in the given example). Subsequently, $N_{\text{gen}}$ samples are drawn for each variable to satisfy the overarching rule $A$. Notably, the image does not depict the discretization of continuous variables or the preservation of inter-variable correlations. However, for illustrative purposes, it is evident that negative values of $x_3$ do not occur under rule $A$, as observed in the synthetic dataset.}
\label{fig:sample_generation}
\end{figure}

To enhance understanding and simplify the process, we present Figure \ref{fig:sample_generation}, which illustrates the data generation steps.

\section{Experiments}

This section outlines the configurations of three anomaly detector models trained on six public datasets and one private dataset from Intesa Sanpaolo (see Table \ref{tab:datasets}), providing explanations using both Anchors and DTOR. The DTOR method and the experiments conducted on public datasets are available in the \texttt{GitHub} repository accessible via the following link: \url{https://github.com/rcrupiISP/DTOR}.

\subsection{Datasets and AD models}

Applying the novel XAI technique across various datasets aims to evaluate its effectiveness in explaining different types of anomalies learned by unsupervised Machine Learning models.

\begin{table}
\centering
\ra{1.3}
\caption{Dataset Details: Each row provides information about a specific dataset, including its identifier (Dataset ID), the number of samples (Num samples), the number of features (Num features), and a brief description (Description). The datasets are sourced from the UCI Machine Learning Repository \cite{Bache+Lichman:2013}.}
\label{tab:datasets}
\resizebox{\textwidth}{!}{
\begin{tabular}{lccr}
\toprule
\textbf{Dataset ID} & \textbf{\# samples} & \textbf{\# features} & \textbf{Description} \\ 
\midrule
\midrule
banking & 100,000 & 26 & \makecell[r]{Banking dataset from Intesa Sanpaolo\\ for identifying anomalies and better analyze\\ the client for possible fraud or criminal behavior.} \\[3ex]
Ionosphere & 351 & 34 & \makecell[r]{Classification of radar returns\\ from the ionosphere.} \\[3ex]
Glass Identification & 214 & 9 & \makecell[r]{From the USA Forensic Science Service,\\ this dataset comprises six types of glass,\\ each defined in terms of their oxide content,\\ including Na, Fe, K, and others.}  \\[3ex] 
lymphography & 148 & 19 & \makecell[r]{This lymphography domain was obtained from\\ the University Medical Centre,\\ Institute of Oncology, Ljubljana, Yugoslavia.} \\[3ex] 
musk v2 & 6,598 & 168 & \makecell[r]{The goal is to learn to predict whether\\ new molecules will be musks or non-musks.} \\[3ex] 
breast cancer wisconsin diagnostic & 198 & 33 & \makecell[r]{Prognostic Wisconsin Breast Cancer Database.} \\[3ex] 
arrhythmia & 452 & 279 & \makecell[r]{Distinguish between the presence and absence\\ of cardiac arrhythmia and classify it into\\ one of the 16 groups.} \\ 
\bottomrule
\end{tabular}
}
\end{table}

The chosen anomaly detector models include IF, One-class SVM and GMM \cite{pedregosa2011scikit}. Default parameters were opted for, as the primary objective of this study is to comprehend the explanation rather than optimize a performance metric specific to the dataset problem. Therefore, three distinct models were chosen to reason in different ways (see Section \ref{sec:intro_ad}). 

The dataset was partitioned into training and testing sets. Specifically, the test set comprises 50 samples from each dataset, containing both anomalies and normal data points. The anomalies for GMM are defined to represent 5\% of the training set, as well as for the isolation forest using the \textit{contamination} hyperparameter set to 0.05. Default hyperparameters were retained for the SVM (kernel: radial basis function, $\nu$ = 0.5, representing the upper bound on the fraction of training errors), resulting in anomalies representing about 50\% of the training set.

\begin{figure}[!hbt]
\centering
\includegraphics[width=1\textwidth]{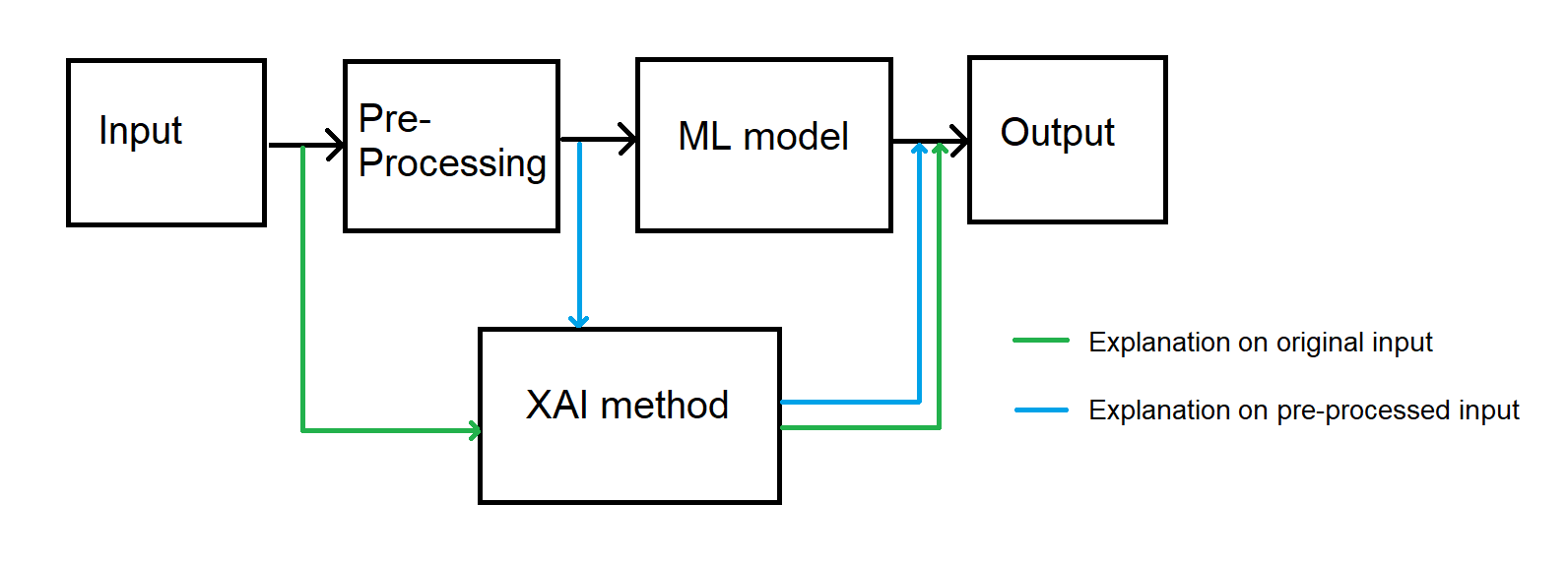}
\caption{Illustration of a machine learning application where the XAI method can provide explanations either in the original input space or the pre-processed one. If the latter option is chosen, the explanation must be converted back into the original feature space, particularly when a rule-based explanation is expected.}
\label{fig:XAI_where_input}
\end{figure}

As part of the pre-processing to prevent convergence issues with the GMM algorithm, the data were standardized. While this may not pose a problem for agnostic XAI techniques, as any input transformation is encapsulated within the ``black box" model, the choice of where to apply the explanation becomes crucial. We opted to apply the explanation directly on the original input space for ease of interpretation, as there is no need to convert the explanation back to the original feature space (see Figure \ref{fig:XAI_where_input}). Additionally, Decision Trees are known to perform well with non-standardized data and are less affected by the curse of dimensionality (i.e., having many input features).

\subsection{Rule-based XAI}

Experiments were conducted using explainability techniques that offer rule-based explanations, as feature importance methods such as SHAP and DIFFI (for the Isolation Forest) proved challenging to interpret, particularly in scenarios with a high number of features. Anchors were initially employed to explain the banking dataset; however, we encountered limitations. Primarily, Anchors can only reason on classification tasks, although the model additionally provides an anomaly score, which can be utilized by the surrogate model. Other limitations pertain to the algorithm itself and its implementation, which were found unsuitable for our specific use case, prompting the development of DTOR.

In addition to Anchors and DTOR, other techniques were considered in this context, such as LORE and RuleXAI. However, LORE is not as user-friendly as Anchors (which can be installed with a simple pip command), requiring more extensive hyperparameter tuning. Additionally, RuleXAI has not been maintained, and some of its Python library requirements have become outdated.

Comparing two explainability techniques is challenging due to the absence of universal metrics to determine superiority, as these metrics depend on the task and the intended audience for the explanations. In our scenario, we approach the problem with the perspective of serving rule-based explanations to Data Scientists, who are often domain experts. Despite this challenge, we summarize the experiments in Table \ref{tab:result}, presenting six informative metrics:

\begin{enumerate}
    \item \textit{Execution time}: the average time taken to generate an explanation rule.
    \item \textit{Precision}: the precision of the rule as defined in eq.~\eqref{eq:precision}.
    \item \textit{Coverage}: the fraction of data points that meet the explanation rule (see eq.~\eqref{eq:coverage}).
    \item \textit{Validity}: the percentage of data points to be explained that effectively satisfy the found rule (see eq.~\eqref{eq:validity}).
    \item \textit{Rule length}: the average length of the rule in terms of its predicates.
\end{enumerate}

It's worth noting the interrelated nature of these metrics: achieving higher coverage may lead to a reduction in precision, while longer rules (i.e. with conditions on more features) may offer more precise explanations but with lower coverage, potentially resulting in reduced interpretability from a human perspective.

Selecting hyperparameters for Anchors and DTOR to ensure fair experimental conditions is also challenging. We opted for parameters that yielded the best results for the banking dataset in terms of both quantity and quality of explanations. However, a dataset-specific approach is necessary to identify the optimal anomaly detector and evaluate quality of explanations effectively.

For Anchors, the precision threshold was set to at least 0.5, which may seem low but is adequate for unbalanced datasets (e.g., with 5\% anomalies). For DTOR, we set the \texttt{max depth} to 8, the \texttt{min impurity decrease} to $10^{-5}$, and the weight $\beta$ for learning the rule to $0.1*N$, where $N$ represents the dataset size. 

It's important to note that DTOR estimates the score rather than the binary output of anomaly/non-anomaly. To obtain this output, the same threshold cut employed by the anomaly detection models is provided. While not reported in this work, each rule in the output can provide not only the precision score but also the average anomaly score, which is more informative. Example rule explanations are presented in Table \ref{tab:examples}, where each rule is associated by its precision, coverage scores and execution time. The comprehensive comparison across multiple datasets and three anomaly detectors is reported in Table \ref{tab:result}.

\begin{table}[htbp]
    \centering
    \caption{Illustrative examples of Isolation Forest explanation on the Lymphography Dataset. The table showcases three dataset samples along with the rules derived from Anchors and DTOR. Each sample is denoted by its corresponding dataset row index, indicating its position within the dataset. While some explanations exhibit similarities, others differ; notably, Anchors often fail to provide any explanation, as observed in example ID 0.}
    \label{tab:examples}
    \begin{tabular}{p{1.2cm}p{5.5cm}p{6cm}}
        \toprule
        \textbf{Index row sample} & \textbf{Anchors Rule} & \textbf{DTOR Rule} \\ 
        \midrule
        36 & \texttt{feature\_9 > 1 AND feature\_7 > 1} & \texttt{feature\_9 > 2 AND feature\_7 > 1.5} \\
        139 & \texttt{feature\_8 <= 1 AND feature\_7 > 1} & \texttt{feature\_0 > 3.5 AND feature\_12 <= 3} \\
        0 & - & \texttt{feature\_4 <= 1.5 AND feature\_15 <= 1.5 AND feature\_10 > 1.5 AND feature\_13 > 3.5} \\
        \bottomrule
    \end{tabular}
\end{table}

\begin{table}[!hbt]
\begin{center}
\caption{Summary of the experiments over seven datasets, three anomaly detectors, and the two compared XAI techniques: Anchors and DTOR. The best performing value between the two methods is highlighted in bold. Execution time refers to the average time (maximum time over the test set in parenthesis). Precision and coverage are reported with standard deviation over the test set. Validation is reported as a percentage. Rule length is the average length over the explanation rules found for the test set.}
\label{tab:result}
\hspace*{-2cm}
\ra{1.2}
\rotatebox{90}{
\begin{tabular}{llcccccc}
\toprule
 &  & \multicolumn{2}{c}{\textbf{Isolation Forest}} & \multicolumn{2}{c}{\textbf{GMM}} & \multicolumn{2}{c}{\textbf{SVM}} \\
 &  & \textbf{Anchors} & \textbf{DTOR} & \textbf{Anchors} & \textbf{DTOR} & \textbf{Anchors} & \textbf{DTOR} \\
\midrule
\midrule
\multirow[c]{5}{*}{banking} & Exec. time & \textbf{13.17 (19.20)} & 24.88 (32.24) & 17.39 (36.86) & \textbf{3.79 (5.98)} & 15.44 (23.75) & \textbf{14.18 (19.58)} \\
 & Precision & 0.54 $\pm$ 0.03 & \textbf{0.90 $\pm$ 0.18} & \textbf{0.64 $\pm$ 0.09} & 0.31 $\pm$ 0.14 & 0.55 $\pm$ 0.04 & \textbf{0.68 $\pm$ 0.26} \\
 & Coverage & 0.00 $\pm$ 0.00 & \textbf{0.15 $\pm$ 0.17} & 0.12 $\pm$ 0.09 & \textbf{0.92 $\pm$ 0.23} & \textbf{0.80 $\pm$ 0.21} & 0.30 $\pm$ 0.25 \\
 & Validity \% & 6 & \textbf{100} & 6 & \textbf{100} & \textbf{100} & \textbf{100} \\
 & Rule length & 0.40 & 6.14 & 0.16 & 8.00 & 1.00 & 8.00 \\
\cline{1-8}
\multirow[c]{5}{*}{glass} & Exec. time & 14.75 (20.94) & \textbf{3.64 (5.13)} & \textbf{0.13 (0.15)} & 2.21 (3.51) & \textbf{0.76 (1.19)} & 2.49 (3.92) \\
 & Precision & 0.16 $\pm$ 0.08 & \textbf{0.89 $\pm$ 0.20} & \textbf{0.74 $\pm$ 0.03} & 0.17 $\pm$ 0.19 & 0.61 $\pm$ 0.10 & \textbf{0.74 $\pm$ 0.20} \\
 & Coverage & 0.01 $\pm$ 0.01 & \textbf{0.10 $\pm$ 0.12} & 0.26 $\pm$ 0.00 & \textbf{0.33 $\pm$ 0.28} & \textbf{0.53 $\pm$ 0.23} & 0.14 $\pm$ 0.14 \\
 & Validity \% & 6 & \textbf{100} & 4 & \textbf{100} & \textbf{100} & \textbf{100} \\
 & Rule length & 0.26 & 5.42 & 0.04 & 7.70 & 1.00 & 6.96 \\
\cline{1-8}
\multirow[c]{5}{*}{ionosphere} & Exec. time & 169.40 (275.22) & \textbf{11.25 (13.61)} & 15.53 (28.24) & \textbf{3.19 (5.03)} & 5.76 (8.03) & \textbf{3.85 (6.62)} \\
 & Precision & 0.73 $\pm$ 0.11 & \textbf{0.91 $\pm$ 0.21} & 0.64 $\pm$ 0.04 & \textbf{0.86 $\pm$ 0.27} & 0.54 $\pm$ 0.04 & \textbf{0.61 $\pm$ 0.23} \\
 & Coverage & 0.00 $\pm$ 0.00 & \textbf{0.08 $\pm$ 0.07} & 0.02 $\pm$ 0.01 & \textbf{0.18 $\pm$ 0.17} & \textbf{0.62 $\pm$ 0.31} & 0.06 $\pm$ 0.07 \\
 & Validity \% & 8 & \textbf{100} & 12 & \textbf{100} & \textbf{100} & \textbf{100} \\
 & Rule length & 0.90 & 6.20 & 0.40 & 7.56 & 1.00 & 7.74 \\
\cline{1-8}
\multirow[c]{5}{*}{lymphography} & Exec. time & 16.94 (40.04) & \textbf{3.90 (5.72)} & 4.20 (7.98) & \textbf{2.36 (3.72)} & \textbf{1.07 (1.35)} & 2.70 (4.40) \\
 & Precision & 0.51 $\pm$ 0.37 & \textbf{0.89 $\pm$ 0.20} & 0.77 $\pm$ 0.26 & \textbf{0.83 $\pm$ 0.15} & 0.56 $\pm$ 0.03 & \textbf{0.70 $\pm$ 0.22} \\
 & Coverage & 0.01 $\pm$ 0.01 & \textbf{0.08 $\pm$ 0.07} & 0.05 $\pm$ 0.03 & \textbf{0.07 $\pm$ 0.08} & \textbf{0.68 $\pm$ 0.23} & 0.03 $\pm$ 0.03 \\
 & Validity \% & 8 & \textbf{100} & 8 & \textbf{100} & \textbf{100} & \textbf{100} \\
 & Rule length & 0.54 & 6.50 & 0.28 & 7.06 & 1.00 & 7.10 \\
\cline{1-8}
\multirow[c]{5}{*}{musk} & Exec. time & 5273.15 (5273.15) & \textbf{11.92 (16.45)} & \textbf{4.97 (5.19)} & 11.05 (18.60) & 472.87 (1374.55) & \textbf{15.54 (27.80)} \\
 & Precision & \textbf{0.99 $\pm$ 0.00} & 0.92 $\pm$ 0.20 & \textbf{0.97 $\pm$ 0.02} & 0.21 $\pm$ 0.26 & \textbf{0.94 $\pm$ 0.03} & 0.87 $\pm$ 0.15 \\
 & Coverage & 0.01 $\pm$ 0.00 & \textbf{0.18 $\pm$ 0.11} & 0.03 $\pm$ 0.02 & \textbf{0.10 $\pm$ 0.12} & \textbf{0.27 $\pm$ 0.16} & 0.05 $\pm$ 0.08 \\
 & Validity \% & 2 & \textbf{100} & 8 & \textbf{100} & \textbf{100} & \textbf{100} \\
 & Rule length & 0.42 & 3.78 & 0.16 & 7.82 & 1.72 & 7.96 \\
\cline{1-8}
\multirow[c]{5}{*}{breast cancer} & Exec. time & 108.08 (108.08) & \textbf{4.49 (6.78)} & \textbf{0.50 (0.50)} & 2.85 (4.65) & 4.80 (9.64) & \textbf{3.67 (6.16)} \\
 & Precision & 0.54 $\pm$ 0.00 & \textbf{0.96 $\pm$ 0.13} & \textbf{0.73 $\pm$ 0.00} & 0.30 $\pm$ 0.34 & 0.59 $\pm$ 0.07 & \textbf{0.86 $\pm$ 0.18} \\
 & Coverage & 0.02 $\pm$ 0.00 & \textbf{0.12 $\pm$ 0.10} & \textbf{0.25 $\pm$ 0.00} & \textbf{0.25 $\pm$ 0.21} & \textbf{0.63 $\pm$ 0.15} & 0.06 $\pm$ 0.10 \\
 & Validity \% & 2 & \textbf{100} & 2 & \textbf{100} & \textbf{100} & \textbf{100} \\
 & Rule length & 0.20 & 6.62 & 0.02 & 7.98 & 1.00 & 7.90 \\
\cline{1-8}
\multirow[c]{5}{*}{arrhythmia} & Exec. time & - & 15.98 (26.03) & - & 9.70 (17.18) & - & 13.19 (24.46) \\
 & Precision & - & 0.91 $\pm$ 0.26 & - & 0.65 $\pm$ 0.44 & - & 0.54 $\pm$ 0.34 \\
 & Coverage & - & 0.23 $\pm$ 0.22 & - & 0.02 $\pm$ 0.06 & - & 0.49 $\pm$ 0.32 \\
 & Validity \% & - & 100 & - & 100 & - & 100 \\
 & Rule length & - & 6.32 & - & 3.70 & - & 7.80 \\
\bottomrule
\end{tabular}
}
\end{center}
\end{table}

\section{Discussion and conclusion}

\autoref{tab:result} shows that DTOR performs at least as good as Anchors, better for certain dataset and metrics. Firstly, it is notable that Anchors often fail to provide an explanation output, with particularly low validity scores, especially for the Isolation Forest (IF) and Gaussian Mixture Model (GMM).\footnote{In some cases, Anchors identify a rule for an instance that does not actually meet the criteria, for instance in the lymphography dataset with the GMM model.} This issue is even more pronounced for the arrhythmia dataset, where the execution encountered out-of-memory errors, preventing the calculation of metrics.

On the other hand, DTOR consistently demonstrates quicker rule discovery times. Additionally, for the Isolation Forest, there is a noticeable preference for DTOR, which is understandable given that IF aims to isolate anomalies through feature space cuts, aligning with DTOR's approach to isolating the sample to be explained.

For the GMM, Anchors exhibit validity below 12\%, and DTOR yields also higher coverage. In contrast, for SVM, there is a trade-off between precision for DTOR and coverage for Anchors, although this could potentially be mitigated by enforcing a shorter rule length in DTOR (e.g., by reducing the \texttt{max depth}). 

Upon closer examination, it became apparent that Anchors are significantly more inclined to provide an explanation for anomalous datapoint rather than ``normal'' points, particularly in unbalanced datasets. In balanced datasets such as those processed by the SVM, both Anchors and DTOR exhibited 100\% validity, since the $\nu$ is kept at the default value of 0.5. The absence of Anchors' rule for normal points could stem from the excessive simplicity of the potential rule, where having no rule already meets the required precision. Additionally, DTOR produced similar rules for normal points, suggesting a consistent behavior and providing insights into defining normal behavior.

Another intriguing observation is that the rules generated for anomalies were often concise, aligning with the simplicity of isolating an anomaly and the Isolation Forest procedure. However, this wasn't consistently observed in explanations for the other two anomaly detector models.

It is important to highlight that this comparison was conducted against a surrogate classifier model, whereas our contribution introduces a surrogate regressor model. In addition to the precision of the rule, DTOR can provide the average anomaly score of a rule, which can offer enhanced interpretability and informativeness. Regarding the computation of precision of rule-based approaches, we introduced a novel methodology for generating the ``neighourhood'', with the aim of retaining some of the correlation between variables that are predicates of the explaining rule, and variables that are not. Incidentally, notice that ---borrowing the terminology of \cite{kumar2020problems}--- this goes in the direction of having a more ``conditional'' type of estimation compared to the ``interventional'' approach of Anchors, whose neighbourhood generation breaks correlations of the data-generating distribution.

In conclusion, DTOR emerges as a simple yet effective technique for addressing XAI in anomaly detection models, particularly from a regression task perspective. It demonstrates versatility across various datasets, delivering comparable or even superior precision and coverage compared to state-of-the-art rule-based XAI algorithms such as Anchors.

%
%
%
\bibliographystyle{splncs04} 
\bibliography{references}

\end{document}